\documentclass{vgtc}                          % final (conference style)
\ifpdf%                                % if we use pdflatex
  \pdfoutput=1\relax                   % create PDFs from pdfLaTeX
  \usepackage{graphicx}                % allow us to embed graphics files
  \DeclareGraphicsExtensions{.pdf,.png,.jpg,.jpeg} % for pdflatex we expect .pdf, .png, or .jpg files
\else%                                 % else we use pure latex
  \usepackage{graphicx}                % allow us to embed graphics files
  \DeclareGraphicsExtensions{.eps}     % for pure latex we expect eps files
\fi%

\usepackage[absolute,overlay]{textpos}
\usepackage{pbox}
\usepackage{everypage}

\newcommand{\stamp}[1][© 2023 IEEE. This is the author's version of the article that has been published in Proceedings of the 2023 IEEE 16th Pacific Visualization Symposium (PacificVis). The final version of this record is available at: \href{https://doi.org/10.1109/PacificVis56936.2023.00030}{\color{blue}10.1109/PacificVis56936.2023.00030}]{%
\begin{textblock*}{140mm}(37mm,270mm)
\centering%
\small% 
\emph{#1}%
\end{textblock*}%
}

\AddEverypageHook{
  \stamp
}

\graphicspath{{figures/}{pictures/}{images/}{./}} % where to search for the images

\usepackage{microtype}                 % use micro-typography (slightly more compact, better to read)
\PassOptionsToPackage{warn}{textcomp}  % to address font issues with \textrightarrow
\usepackage{textcomp}                  % use better special symbols
\usepackage{mathptmx}                  % use matching math font
\usepackage{times}                     % we use Times as the main font
\usepackage{cite}                      % needed to automatically sort the references
\usepackage{tabu}                      % only used for the table example
\usepackage{booktabs}                  % only used for the table example
\makeatletter
\renewcommand{\ps@plain}{%
\renewcommand{\@oddfoot}{\hfil\textrm{\thepage}\hfil}%
\renewcommand{\@evenfoot}{\@oddfoot}%
}
\renewcommand{\ps@empty}{%
\renewcommand{\@oddfoot}{\hfil\textrm{\thepage}\hfil}%
\renewcommand{\@evenfoot}{\@oddfoot}%
}
\makeatother

\usepackage[dvipsnames]{xcolor} % Remove that in the end!
\definecolor{blue}{RGB}{0, 112, 192}
\definecolor{red}{RGB}{192, 0, 0}

\onlineid{0}

\vgtccategory{Research}

\vgtcinsertpkg

\title{MetaStackVis: Visually-Assisted Performance Evaluation of Metamodels}

\author{Ilya Ploshchik\thanks{e-mail: ip222gs@student.lnu.se}\\ %
        \scriptsize Linnaeus University %
\and Angelos Chatzimparmpas\thanks{e-mail: angelos.chatzimparmpas@lnu.se}\\ %
     \scriptsize Linnaeus University
\and Andreas Kerren\thanks{e-mail:  andreas.kerren\{@lnu.se, @liu.se\}}\\ %
     \parbox{1.4in}{\scriptsize \centering Linnaeus University \\ Linköping University}}
     
\teaser{
    \vspace{-5.5mm}
    \centering
    \includegraphics[width=\linewidth]{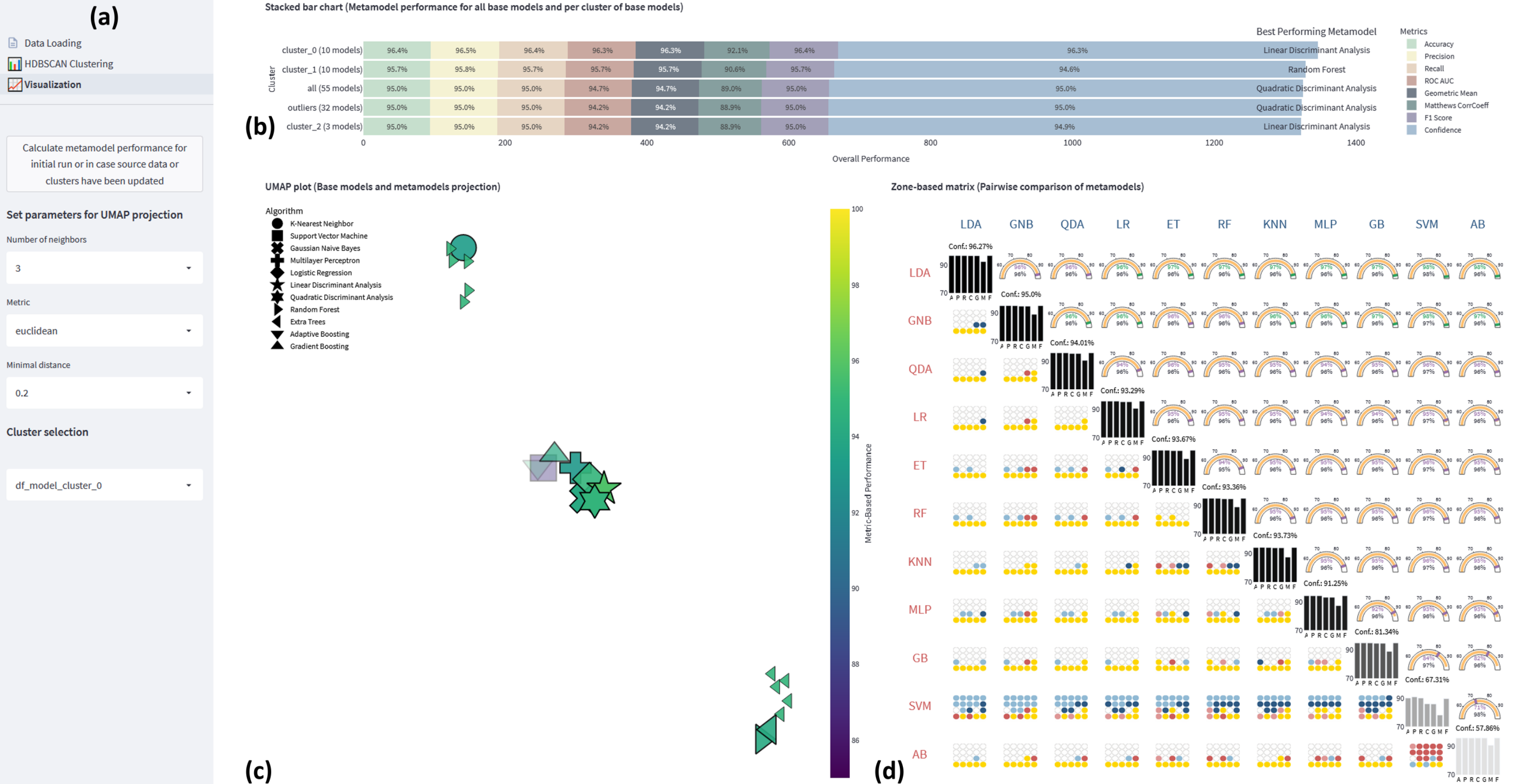}\vspace{-4mm}
    \caption{Comparing alternative metamodels' predictive performance with MetaStackVis: (a) a panel for the selection of UMAP hyperparameters and the active cluster; (b) a stacked bar chart for identifying the best-performing metamodel for each cluster of base models by analyzing validation metrics and confidence scores; (c) a UMAP plot that aggregates the results of the predicted probabilities and the metric-based performance for the base models and metamodels of the active cluster; and (d) a zone-based matrix that combines all pairs of metamodels to designate the possible benefits of an extra stacking layer due to the juxtaposition of the soft voting outcome (cf. grid of points) versus the optimal outcome from the merger between two metamodels (see gauge charts).}
    \label{fig:system}
}

\abstract{
Stacking (or stacked generalization) is an ensemble learning method with one main distinctiveness from the rest: even though several base models are trained on the original data set, their predictions are further used as input data for one or more metamodels arranged in at least one extra layer. Composing a stack of models can produce high-performance outcomes, but it usually involves a trial-and-error process. Therefore, our previously developed visual analytics system, StackGenVis, was mainly designed to assist users in choosing a set of top-performing and diverse models by measuring their predictive performance. However, it only employs a single logistic regression metamodel. In this paper, we investigate the impact of alternative metamodels on the performance of stacking ensembles using a novel visualization tool, called MetaStackVis. Our interactive tool helps users to visually explore different singular and pairs of metamodels according to their predictive probabilities and multiple validation metrics, as well as their ability to predict specific problematic data instances. MetaStackVis was evaluated with a usage scenario based on a medical data set and via expert interviews.
} % end of abstract

\CCScatlist{
  \CCScatTwelve{Human-centered computing}{Visu\-al\-iza\-tion}{Visualization systems and tools}{}
}

\begin{document}

\stamp

\maketitle

%% The ``\maketitle'' command must be the first command after the
%% ``\begin{document}'' command. It prepares and prints the title block.

%% the only exception to this rule is the \firstsection command
\section{Introduction} \label{sec:intro}%
  \noindent Stacking (also called~\emph{stacked generalization}~\cite{Wolpert1992Stacked}) is a machine learning (ML) paradigm that operates with heterogeneous ML models arranged in one or more layers, where each subsequent layer summarizes the previous ML models' predictions~\cite{Sagi2018Ensemble}. A model in such a context represents the output structure of the ML algorithm after fitting it to data and selecting a specific hyperparameter set. In its simplest form, the stacking ensemble is composed of two layers: layer 0, which comprises multiple base models, and layer 1, which contains one or more metamodels~\cite{Chatzimparmpas2021Empirical}. Stacking is a popular method that typically increases the predictive performance due to the deployment of several models. It can also attain a low bias and low variance simultaneously~\cite{Kohavi1996Bias}, especially when juxtaposed against a single ML model~\cite{Ting1997Stacked}. However, one should ensure getting the top-performing and diverse models by carefully choosing ML algorithms on the underlying and metamodel layers~\cite{Naimi2018Stacked}.

% \begin{figure}[tb]
%     \centering
%     \includegraphics[width=\columnwidth]{StackGenVis-crop.pdf}\vspace{-4mm}
%     \caption{StackGenVis allows users to visually explore ML models from various algorithms and select the best-performing and most diversified base models. However, it only supports logistic regression as the metamodel. Our prior publication includes further information~\cite{Chatzimparmpas2021StackGenVis}.}
%     \label{fig:StackGenVis}
%     \vspace{-2mm}
% \end{figure}

To eliminate---as much as possible---trial-and-error processes and manage the complete procedure of building impactful stacking ensembles, interactive visual analytics (VA) solutions have been demonstrated to be very effective~\cite{Chatzimparmpas2021StackGenVis}. Nevertheless, there has been little work to formally investigate further the influence of alternative metamodels on predictive performance (e.g., by visually comparing various metamodels). In a stacking ensemble scenario, Latha and Jeeva~\cite{LATHA2019Improving} found that the random forest algorithm is better in terms of predictive capability when used as a metamodel compared to random trees for the particular healthcare data set they examined. However, the choice of different metamodels depends on the given problem, and the applications of the same metamodels to local subsets of base models instead of all ML models could impact their prediction~\cite{Zhang2022SliceTeller}. Furthermore, the current literature poorly covers the detection of behavioral differences between the most powerful base models' predictions and the metamodels' results. Finally, an indication that an additional layer would be beneficial when composing a stacking ensemble is another interesting but yet unexplored issue.

This short paper presents a follow-up analysis of the impact of different metamodels on the prediction and works as a stand-alone extension tool of our IEEE VAST'20 paper on StackGenVis~\cite{Chatzimparmpas2021StackGenVis}, a \emph{source-available VA system}~\cite{StackGenVisCode} for constructing high-performance stacking ensembles. StackGenVis is the first VA system~\cite{Chatzimparmpas2020The} specially designed to visually monitor and handle the stacking process from scratch with the selection of diverse algorithms and concrete models, including data wrangling support. Its current version was developed using logistic regression with default hyperparameters at the metamodel layer. In this work, we present MetaStackVis (see~\autoref{fig:system}), an \emph{interactive and open-source visualization tool}~\cite{MetaStackVisCode} for exploring alternative metamodels after extracting base models from StackGenVis. In addition to comparing various models' configurations based on multiple validation metrics and predictive probabilities (or \emph{confidence}) on the basic and metamodel levels, we also combined pairs of metamodels and contrasted their overall predictive power, as well as local performance for continuously misclassified data instances.
%\autoref{sec:sys} describes the different views of the tool, while \autoref{sec:use} illustrates the applicability and usefulness of MetaStackVis in a diabetes prediction setting. Next in \autoref{sec:con}, we examine the feedback our tool obtained from interview sessions conducted with ML and VA experts and present the identified limitations that can lead to further improvement of our tool. Finally, \autoref{sec:con} concludes our paper.

\section{MetaStackVis: System Overview} \label{sec:sys}
  MetaStackVis is a visualization tool implemented in Jupyter Notebook~\cite{jupyter} with Plotly~\cite{plotly} as the visualization library and Scikit-Learn~\cite{Pedregosa2011Scikit} for ML purposes. It is then deployed using Streamlit~\cite{streamlit}.

\textbf{Data Loading Tab.} In order to employ this visualization tool, users have to initially experiment with the most recent publicly available version of StackGenVis and extract the predicted probabilities for each data instance and the scores of all validation metrics for the ML models they prefer in the form of two separate CSV files. The latter file with the scores should also contain the hyperparameters used for every ML algorithm to produce the exported ML models. In this section, we explain MetaStackVis with the beginner-friendly Breast Cancer Wisconsin data set~\cite{Dua2017Machine}. The data set includes records for 699 breast cancer cases, labeled as either benign or malignant depending on nine features. Through the StackGenVis system, we split the data into training and testing sets with an 80/20 ratio. For each of the 11 supported ML algorithms, StackGenVis uses pre-defined hyperparameter sets to generate concrete models. In all our experiments, we selected the top 5 base models in terms of overall performance per algorithm, leading to 55 base models in total.

\textbf{HDBSCAN Clustering Tab.} Afterwards, the base models are grouped into separate clusters using HDBSCAN~\cite{Campello2013Density} according to their predicted probabilities for all test instances. HDBSCAN is a state-of-the-art algorithm that focuses on creating groupings with base models that predict similar testing data subsets~\cite{Xu2015Comprehensive}. MetaStackVis allows users to test different hyperparameter combinations for this clustering algorithm (see the first three columns of~\autoref{tab:hdbscan}) while optimizing the solution for the \emph{density-based clustering validation} (DBCV)~\cite{Moulavi2014Density} and the \emph{coverage} scores, concurrently. The former is calculated by computing the density within a cluster and the density between clusters to contribute to the weighted sum of ``validity index'' values of clusters, with higher density within a cluster and lower density between clusters indicating better results than the opposite case. The latter represents how many base models fall into specific clusters; thus, all base models that belong to an existing cluster, except for the \emph{outliers/noisy data}, are divided by the 55 base models. The last three columns of~\autoref{tab:hdbscan} present the results for the top 5 cluster compositions with the best-found DBCV, Coverage, and a combination of both scores. This table is sorted in descending order based on the multiplication of DBCV and coverage. The goal is to minimize the number of outliers due to a high coverage score and keep the cluster formations---as precise as possible---with the DBCV heuristic. However, users can still manually choose the hyperparameters from a pre-determined list of options. MetaStackVis facilitates users with either testing a new hypothesis or picking the \emph{default} cluster division according to the highest \emph{DBCV\_Coverage}.

\begin{table}[t]
    \caption{Different hyperparameter sets are being used for the HDBSCAN clustering to find the top 5 similarly-performing cluster compositions. The table is sorted according to the \emph{DBCV\_Coverage} column.}\vspace{-2mm} 
    \includegraphics[width=\columnwidth]{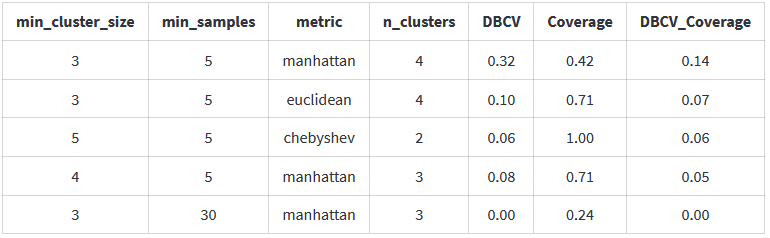}
    \label{tab:hdbscan}
    \vspace{-8mm}
\end{table}

\textbf{Visualization Tab.} After exploring the desired cluster compositions, the \emph{n\_clusters} column defines the number of clusters. The division of base models in them occurs from the HDBSCAN algorithm. We proceed with the defaults, resulting in: \emph{cluster\_0} and \emph{cluster\_1} containing 10 models each, \emph{outliers} (i.e., unassigned base models by HDBSCAN) forming a cluster of 32 models, and \emph{cluster\_2} including 3 models as illustrated in~\autoref{fig:system}(b), in parentheses. Next, the 11 metamodels will be trained upon the base models of all the aforementioned setups. We distill the hyperparameters from every ML algorithm's top-performing base model to use them for the metamodels originating from the same algorithms. However, a plethora of hyperparameter tuning alternatives can be found in the literature~\cite{Shahhosseini2022Optimizing} (cf.~\autoref{sec:eval} for such limitations). The Visualization tab in \autoref{fig:system}(a) incorporates options for selecting UMAP hyperparameters~\cite{McInnes2018UMAP} and the cluster under investigation, as well as three different views: \emph{stacked bar chart}, \emph{UMAP plot}, and \emph{zone-based matrix}.

\begin{figure*}[ht]
    \includegraphics[width=\textwidth]{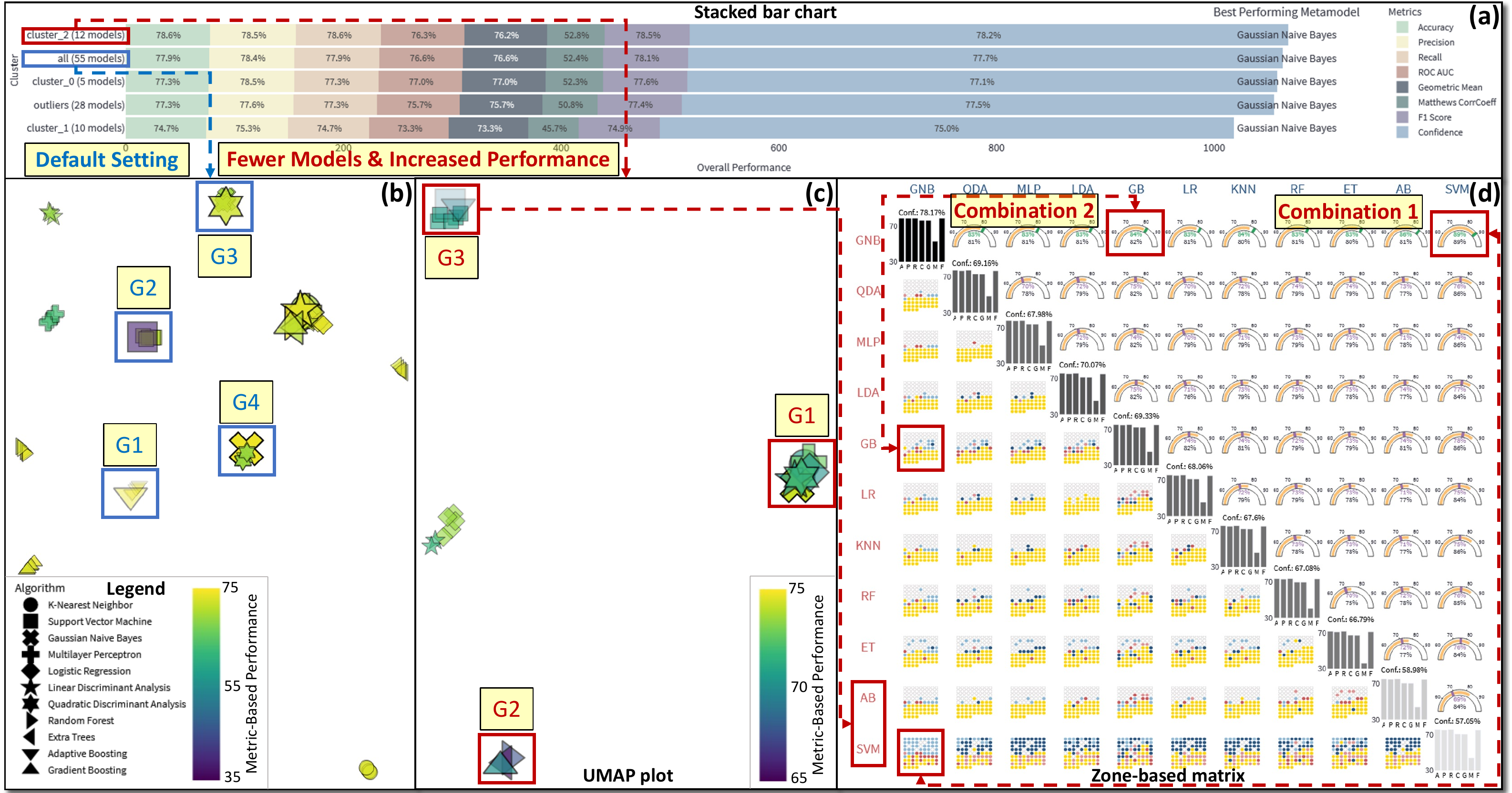}\vspace{-4mm} 
    \caption{The investigation of \emph{all} and \emph{cluster\_2} comprising 12 base models. View (a) presents the performance of the best-performing metamodel for each cluster according to the seven validation metrics and confidence. The UMAP visible in (b) gathers base models and metamodels predicting similarly the same test instances in groups (Gs) such as \textcolor{blue}{G1}-- \textcolor{blue}{G4}. On the other hand, (c) visualizes \emph{cluster\_2}, with \textcolor{red}{G1} showcasing that most of the metamodels perform identically, \textcolor{red}{G2} solely with tree-based ML algorithms, and \textcolor{red}{G3} with the two most unconfident metamodels. The unification of predictions from pairs of diverse metamodels is also possible as seen in (d), leading to two promising combinations.}
    \label{fig:usage}
    \vspace{-3mm}
\end{figure*}

The \textbf{stacked bar chart} in~\autoref{fig:system}(b) presents the best-performing metamodel in each cluster, including all base models and the group of outliers for seven different validation metrics also supported by StackGenVis. This visualization provides an overview of performance (in percentage \% format) for the best candidate from the 11 metamodels created in every cluster, using the following metrics: Accuracy, Precision, Recall, ROC AUC, Geometric Mean, Matthews Correlation Coefficient (CorrCoeff), F1 Score, and Confidence. The last metric is the \emph{average predicted probability} for all test instances. Additionally, we convert Matthews CorrCoeff to an absolute value ranging from 0 to 100\%. The average of all seven validation metrics plus the confidence is then divided by 2 in order to compute the Overall Performance that defines the ranking of the clusters from top to bottom in this visualization (i.e., from best to worst). Therefore, Confidence is multiplied seven times to capture the same space as all validation metrics because users should be able to compare the two main components of overall performance globally. The legend for this view maps the metrics to the different color encodings. If a user deems a metric useless for the given problem, they can deselect this metric and temporarily hide it. If we compare the total length of the stacked bars in~\autoref{fig:system}(b), \emph{cluster\_0} contains only 10 instead of 55 base models and reaches the highest overall performance with Linear Discriminant Analysis as the metamodel.

The \textbf{UMAP plot} in~\autoref{fig:system}(c) enables the visual exploration of the base models belonging to the active cluster selected before and the 11 metamodels summarizing their predictions. Hence, offering a deeper behavioral analysis of all metamodels in contrast to the base models. Each point is one model, with base models being smaller in size, while the opposite is true for the metamodels. The UMAP projects the high-dimensional predicted probabilities calculated for the provided data set into two dimensions. In our example, groups of points represent clusters of models that perform similarly according to 140 test instances (which is the 20\% testing set). A summary of the performance of each model according to the average value computed from the seven validation metrics is designated as Metric-Based Performance in~\autoref{fig:system}(c) and is being color-encoded using the Viridis colormap~\cite{Liu2018Somewhere}. The legend on the left-hand side of this visualization maps the different algorithms as 11 distinguishable symbols for each ML algorithm. For example, the right-pointing arrows are the models constructed from random forest and the left-pointing arrows from extra trees. The opacity of the models is used for the confidence previously introduced, with a higher value forcing the ML model to be more opaque and vice versa.

The \textbf{zone-based matrix} in~\autoref{fig:system}(d) is inspired by the scatterplot matrix~\cite{Carr1987High}, and it provides a more comprehensive perspective of the metamodels' performance. We designed three different zones: the matrix diagonal, the lower triangular part, and the upper triangular part. A bar chart in the matrix diagonal visualizes the metric-based performance of the validation metrics individually as a bar. Color and text convey the confidence (Conf.) of each metamodel, ranked from the best- to the worst-performing one, as already explained for view (b) in~\autoref{fig:system}. Black denotes the highest confidence value, while light gray is the lowest possible. The remaining zones allow users to perform pairwise comparisons between all combinations of metamodels. The lower triangular part demonstrates the union of all misclassified test instances by at least one metamodel pair (20 in our example). The points in the grid are sorted according to the sum of predicted probabilities for all combinations, leading to the easiest-to-classify test samples always being on top (in white, if correctly classified by both metamodels) and the hardest-to-classify at the bottom (in yellow color, if wrongly classified by both metamodels). As a reference model, we apply the soft majority voting strategy~\cite{Cai2013A} (i.e., predicted probabilities being used) with dark red when the row-wise metamodels are unable to overcome the wrong prediction of the blue metamodels and light red in case these metamodels are correct and their confidence surpasses the other metamodel. Thus, more prominent colors highlight the points and demonstrate the failure of metamodels to predict these points correctly. On the contrary, the upper triangular part is about the ``theoretically achievable maximum'' predictive performance if the optimal metamodel was selected for all the test instances (140 in our case). The gauge charts represent the average of all validation metrics' performance in orange (and in the black text below) and the higher or lower confidence value compared to this metric-based performance in green or purple colors, respectively. The exploration of metamodel pairs in~\autoref{fig:system}(d) aims to indicate the available room for other schemas, such as establishing an extra stacking layer to aggregate the predictions of this layer.

\section{Usage Scenario} \label{sec:use}%
  Supposedly Mia is a data scientist working in a hospital. She receives data about 268 positive and 500 negative patients with eight features related to Pima Indian Diabetes~\cite{Smith1988Using}. Her task is to improve predictive performance using stacking ensembles. After exploring the data, ML algorithms, and base models with StackGenVis, she deploys MetaStackVis to experiment with alternative metamodels.

\textbf{Overview.} First, Mia loads the exported data into MetaStackVis and chooses the setting for attaining the highest score for the HDBSCAN clustering algorithm. The hyperparameters are automatically set to 3, 5, and Chebyshev metric for the first three column headers visible in~\autoref{tab:hdbscan}, respectively; thus, resulting in four clusters in total with a DBCV of 0.17 and Coverage of 0.49. All clusters are observable in~\autoref{fig:usage}(a), with \emph{cluster\_2} consisting of fewer models (i.e., 12 base models instead of 55) and at the same time achieving better overall performance for the best-performing metamodel compared to \emph{all}. Mia focuses on the UMAP plots for both settings, using the Manhattan metric because it is preferable for high-dimensionality~\cite{Aggarwal2001On}.

\textbf{Detailed Exploration of Base Models and Metamodels.} An interesting finding when exploring \emph{all} in~\autoref{fig:usage}(b) is that the base models are grouped depending on their origin and, in general, reach similar metric-based performance regardless of the hyperparameters chosen. The same effect can be observed for the metamodels: for example, Group 1 (\textcolor{blue}{G1}) and \textcolor{blue}{G2} are formed because they predict similarly the same test instances. However, Mia is surprised by the fact that the metamodel in \textcolor{blue}{G3} and \textcolor{blue}{G4} have swapped places, with Gaussian Naive Bayes (GNB) being closer to Quadratic Discriminant Analysis and vice versa. This pattern requires a deeper investigation.

\textbf{Pairwise Combination of Metamodels.} Mia continues her exploration with the more efficient and effective \emph{cluster\_2}. From~\autoref{fig:usage}(c), she acknowledges that most of the metamodels belong to \textcolor{red}{G1}. An exception here is \textcolor{red}{G2}, which only contains tree-based ML algorithms such as Gradient Boosting (GB) and \textcolor{red}{G3} with Adaptive Boosting and Support Vector Machine (SVM). The last group's metamodels have the lowest confidence according to~\autoref{fig:usage}(d), but Mia sees the benefit of combining diverse metamodels. One candidate is the combination of GNB and SVM with 89\% theoretical maximum prediction score and confidence (see gauge charts). When combined with a majority voting strategy, the GNB can correctly predict most easy test instances on top, as shown in the grid of points with light blue color, but SVM fails to overcome GNB for the difficult-to-classify test cases at the bottom (due to dark red colors). Another eye-catching combination is GNB and GB, with 82\% hypothetical maximum metric-based performance and 84\% confidence. According to the grid of points, except for the misclassified instances in yellow, four cases are correctly classified by GB and one by GNB but neither can reach higher confidence than its counterpart. Consequently, Mia understands that a potential extra layer summarizing this second layer's predictions could improve the combination of those metamodels.
 % end of In-depth Analysis

%% Discussion section.
\section{Evaluation} \label{sec:eval}%
  We performed online semi-structured interviews with four experts asynchronously to obtain qualitative feedback on the usefulness of MetaStackVis, using the same criteria from prior works~\cite{Xu2019EnsembleLens,Ma2020Explaining,Chatzimparmpas2021Visevol,Chatzimparmpas2022FeatureEnVi}. 

\textbf{Participants.} The first ML expert (\textbf{E1}) is a senior lecturer in mathematics with a PhD in this field and has 4.5 years of experience with ML. The second ML expert (\textbf{E2}) is an assistant professor focusing on ML and deep learning with 7.5 years of experience in ML. The third VA expert (\textbf{E3}) is a senior lecturer working with clustering and dimensionality reduction, and he has 6 years of experience with ML. The fourth VA expert (\textbf{E4}) is a senior lecturer focusing on natural language processing and applied ML with approximately 10 years of experience. The first three experts have reported no colorblindness issues. Although \textbf{E4} has a mild case of colorblindness, he affirmed having no difficulty accurately recognizing the specific color combinations we utilized in the MetaStackVis implementation.

\textbf{Methodology.} Each interview lasted about one hour, and the interviews were conducted as follows: (1) an introduction to our visualization tool's main goals; (2) a display of the functionality of each visualization and interaction with the tool using the Pima Indian Diabetes data set; and (3) a discussion of the processes followed to arrive at the findings in~\autoref{sec:use}. We asked the participants to comment on anything. Their major points are summarized below.

\textbf{Workflow.} \textbf{E2} and \textbf{E4} mentioned that the overall proposed workflow makes sense and is appropriately reflected in the spatial arrangement of the views. In particular, \textbf{E2} commented on the progressive analysis of the different base model clusters and the produced metamodels as a positive aspect of MetaStackVis. However, \textbf{E1} and \textbf{E2} suggested a different approach for selecting clusters: concentrate first on the UMAP plot and let users pick clusters of interest with different base models, but they both admitted that using a user-controlled HDBSCAN clustering should be a practical starting point before the manual cluster exploration. \textbf{E1} stated that the Meta\-StackVis workflow requires a thorough understanding of stacking ensembles and access to the publicly-available StackGenVis source code~\cite{StackGenVisCode}. Since it may also be advantageous to feed the generated human knowledge from MetaStackVis back to StackGenVis, as \textbf{E4} framed it, we intend to unify all features in a ``single tool'' solution.

\textbf{Visualization and Interaction.} \textbf{E3} had an overall good impression regarding the choice of visual representations to map the computed data. \textbf{E1} was amazed by the details provided in MetaStackVis but admitted that it could be slightly overwhelming when one sees the tool for the first time. Specifically, \textbf{E2} and \textbf{E3} mentioned that the stacked bar chart and the UMAP plot are easy to interpret, but the zone-based matrix can be challenging to grasp and certainly needs additional time to understand. An improvement here could be to highlight the left-hand side with all confusing data points for the combinations of the 11 metamodels and the right-hand side with the gauge charts when users hover over either one of them because it would become easier to perform pairwise comparisons of the fused metamodels, as \textbf{E2} pointed out. Nevertheless, \textbf{E3} supported the idea that ordering metamodels from the best- to the worst-performing one helps interpret the zone-based matrix. Here, \textbf{E3} and \textbf{E4} mentioned that the better-performing pairs of metamodels could be considered future candidates for input to a potential third layer of metamodels. \textbf{E3} was fascinated by the clusters visible in the UMAP plot, which translates to the prediction capability of each model in the test data set. \textbf{E3} mentioned that this visualization illustrates our successful extraction of useful information and patterns related to cluster structure. He then continued: ``it would be complicated to transform that information into reliable insights about the performance without support from the other views''. \textbf{E2} agreed that the UMAP plot could suggest how differently the base models perform compared to the metamodels. \textbf{E3} proposed to segmentize and vertically align each metric instead of the global sorting, but he understands that will affect the global ranking. This observation partially matches with \textbf{E1}'s comment to focus on one or a group of validation metrics at a time and not have all seven visualized simultaneously. Although MetaStackVis already allows users to hide irrelevant metrics, this feature should be implemented in the future (as with StackGenVis).

\textbf{Limitations.} \emph{Efficiency} and \emph{scalability} were two concerns raised by \textbf{E4}. The former refers to the required computation time to render all views. However, this does not threaten interactivity as long as everything gets parallelized and/or pre-computed beforehand~\cite{Li2020P}. For the latter case, he pointed out the tool's limitation to visualize a much larger data set with more difficult-to-predict instances due to the increased space demand for the zone-based matrix. A simple solution to this problem could be filtering, which applies to scenarios where some metamodel pairs are performing poorly. As \textbf{E2} stated, the tool works solely with \emph{binary classification problems} and does not support \emph{alternative hyperparameter optimization techniques}~\cite{Feurer2019Hyperparameter}. \textbf{E1} referred to the important role that metamodels' confidence plays in the data exposition, but instead of being aggregated as in our tool, it could be beneficial to use \emph{individual visual representations of spread}. He continued to say that it is necessary to visualize the data distribution on demand to better relate to the underlying \emph{explanation of why some instances are constantly misclassified}. In the future, we plan to improve MetaStackVis to overcome such limitations.

%\textbf{Participants.} The first ML expert (\textbf{E1}) is a senior lecturer in mathematics with a PhD in this field and has 4.5 years of experience with ML. He also has basic knowledge regarding ensemble learning and currently works with reinforcement learning. The second ML expert (\textbf{E2}) is an assistant professor focusing on ML and deep learning with 7.5 years of experience in ML. His PhD is in media technology, and he has worked with ensemble learning methods for a few years. The third VA expert (\textbf{E3}) is a senior lecturer working with clustering and dimensionality reduction, and he has 6 years of experience with ML and 2 years of experience with ensemble learning. The fourth VA expert (\textbf{E4}) is a senior lecturer focusing on natural language processing and applied ML with approximately 10 years of experience. He has 2 years of experience with ensemble learning, in particular. The last two experts have completed their PhDs within the information visualization and VA disciplines. The first three experts have reported no colorblindness issues. Although \textbf{E4} has a mild case of colorblindness, he affirmed having no difficulty accurately recognizing the specific color combinations we utilized in the MetaStackVis implementation.

\section{Conclusion} \label{sec:con}%
  \noindent In this paper, we presented MetaStackVis, a visualization tool that enables users to visually assess the performance of metamodels in stacking ensemble learning. It allows users to tune HDBSCAN and apply metamodels to different cluster compositions of base models. Users can also compare the metamodels based on seven validation metrics and their average predicted probability, observe the performance similarities with the underlying base models, and check for powerful pairwise combinations of metamodels that hint at the possible benefit of introducing an extra stacking layer. The applicability and effectiveness of MetaStackVis were evaluated using a real-world healthcare data set and interviews with four experts, who suggested that the comparison of alternative metamodels with our tool is promising. Finally, they helped us recognize the current limits of MetaStackVis, which we will work on in the future.

%% if specified like this the section will be committed in review mode
\acknowledgments{
This work was partially supported through the ELLIIT environment for strategic research in Sweden.}

\bibliographystyle{abbrv-doi}

\bibliography{MetaStackVis}

\begin{thebibliography}{10}

\bibitem{Aggarwal2001On}
C.~C. Aggarwal, A.~Hinneburg, and D.~A. Keim.
\newblock On the surprising behavior of distance metrics in high dimensional
  space.
\newblock In {\em Proceedings of the International Conference on Database
  Theory (ICDT)}, pp. 420--434. Springer Berlin Heidelberg, 2001.

\bibitem{Cai2013A}
J.~Cai, J.~L. Garner, and R.~A. Walkling.
\newblock A paper tiger? {A}n empirical analysis of majority voting.
\newblock {\em Journal of Corporate Finance}, 21:119--135, 2013. doi: {{%
10\hspace{.1pt}\discretionary{.}{%
}{.}\hspace{.4pt}1016\discretionary{/}{%
}{/}j\hspace{.1pt}\discretionary{.}{%
}{.}\hspace{.4pt}jcorpfin\hspace{.1pt}\discretionary{.}{%
}{.}\hspace{.4pt}2013\hspace{.1pt}\discretionary{.}{%
}{.}\hspace{.4pt}01\hspace{.1pt}\discretionary{.}{%
}{.}\hspace{.4pt}002}}


\bibitem{Campello2013Density}
R.~J. Campello, D.~Moulavi, and J.~Sander.
\newblock Density-based clustering based on hierarchical density estimates.
\newblock In {\em Proceedings of the Pacific-Asia Conference on Knowledge
  Discovery and Data Mining}, pp. 160--172. Springer, 2013.

\bibitem{Carr1987High}
D.~B. Carr, R.~J. Littlefield, W.~L. Nicholson, and J.~S. Littlefield.
\newblock Scatterplot matrix techniques for large {N}.
\newblock {\em Journal of the American Statistical Association},
  82(398):424--436, 1987.

\bibitem{Chatzimparmpas2020The}
A.~Chatzimparmpas, R.~M. Martins, I.~Jusufi, K.~Kucher, F.~Rossi, and
  A.~Kerren.
\newblock The state of the art in enhancing trust in machine learning models
  with the use of visualizations.
\newblock {\em Computer Graphics Forum}, 39(3):713--756, June 2020. doi: {{%
10\hspace{.1pt}\discretionary{.}{%
}{.}\hspace{.4pt}1111\discretionary{/}{%
}{/}cgf\hspace{.1pt}\discretionary{.}{%
}{.}\hspace{.4pt}14034}}


\bibitem{Chatzimparmpas2021Empirical}
A.~Chatzimparmpas, R.~M. Martins, K.~Kucher, and A.~Kerren.
\newblock Empirical study: {V}isual analytics for comparing stacking to
  blending ensemble learning.
\newblock In {\em Proceedings of the 23rd International Conference on Control
  Systems and Computer Science (CSCS)}, pp. 1--8. IEEE, 2021.

\bibitem{Chatzimparmpas2021StackGenVis}
A.~Chatzimparmpas, R.~M. Martins, K.~Kucher, and A.~Kerren.
\newblock {StackGenVis}: {A}lignment of data, algorithms, and models for
  stacking ensemble learning using performance metrics.
\newblock {\em IEEE Transactions on Visualization and Computer Graphics},
  27(2):1547--1557, Feb. 2021. doi: {{%
10\hspace{.1pt}\discretionary{.}{%
}{.}\hspace{.4pt}1109\discretionary{/}{%
}{/}TVCG\hspace{.1pt}\discretionary{.}{%
}{.}\hspace{.4pt}2020\hspace{.1pt}\discretionary{.}{%
}{.}\hspace{.4pt}3030352}}


\bibitem{Chatzimparmpas2021Visevol}
A.~Chatzimparmpas, R.~M. Martins, K.~Kucher, and A.~Kerren.
\newblock {VisEvol}: {V}isual analytics to support hyperparameter search
  through evolutionary optimization.
\newblock {\em Computer Graphics Forum}, 40(3):201--214, June 2021. doi: {{%
10\hspace{.1pt}\discretionary{.}{%
}{.}\hspace{.4pt}1111\discretionary{/}{%
}{/}cgf\hspace{.1pt}\discretionary{.}{%
}{.}\hspace{.4pt}14300}}


\bibitem{Chatzimparmpas2022FeatureEnVi}
A.~Chatzimparmpas, R.~M. Martins, K.~Kucher, and A.~Kerren.
\newblock {FeatureEnVi}: {V}isual analytics for feature engineering using
  stepwise selection and semi-automatic extraction approaches.
\newblock {\em IEEE Transactions on Visualization and Computer Graphics},
  28(4):1773--1791, 2022. doi: {{%
10\hspace{.1pt}\discretionary{.}{%
}{.}\hspace{.4pt}1109\discretionary{/}{%
}{/}TVCG\hspace{.1pt}\discretionary{.}{%
}{.}\hspace{.4pt}2022\hspace{.1pt}\discretionary{.}{%
}{.}\hspace{.4pt}3141040}}


\bibitem{Dua2017Machine}
D.~Dua and C.~Graff.
\newblock {{UCI} Machine Learning Repository}, 2017.

\bibitem{Feurer2019Hyperparameter}
M.~Feurer and F.~Hutter.
\newblock Hyperparameter optimization.
\newblock In {\em Automated Machine Learning: Methods, Systems, Challenges},
  pp. 3--33. Springer International Publishing, 2019. doi: {{%
10\hspace{.1pt}\discretionary{.}{%
}{.}\hspace{.4pt}1007\discretionary{/}{%
}{/}978\discretionary{%
}{-}{-}3\discretionary{%
}{-}{-}030\discretionary{%
}{-}{-}05318\discretionary{%
}{-}{-}5\_1}}


\bibitem{jupyter}
{Jupyter} --- {A} web-based interactive computing platform.
\newblock \url{https://jupyter.org}, 2014.
\newblock Accessed December 10, 2022.

\bibitem{Kohavi1996Bias}
R.~Kohavi and D.~Wolpert.
\newblock Bias plus variance decomposition for zero-one loss functions.
\newblock In {\em Proceedings of the International Conference on Machine
  Learning}, ICML~'96, pp. 275--283. Morgan Kaufmann Publishers Inc., 1996.

\bibitem{LATHA2019Improving}
C.~B.~C. Latha and S.~C. Jeeva.
\newblock Improving the accuracy of prediction of heart disease risk based on
  ensemble classification techniques.
\newblock {\em Informatics in Medicine Unlocked}, 16:100203, 2019. doi: {{%
10\hspace{.1pt}\discretionary{.}{%
}{.}\hspace{.4pt}1016\discretionary{/}{%
}{/}j\hspace{.1pt}\discretionary{.}{%
}{.}\hspace{.4pt}imu\hspace{.1pt}\discretionary{.}{%
}{.}\hspace{.4pt}2019\hspace{.1pt}\discretionary{.}{%
}{.}\hspace{.4pt}100203}}


\bibitem{Li2020P}
J.~K. Li and K.-L. Ma.
\newblock {P4}: {P}ortable parallel processing pipelines for interactive
  information visualization.
\newblock {\em IEEE Transactions on Visualization and Computer Graphics},
  26(3):1548--1561, 2020. doi: {{%
10\hspace{.1pt}\discretionary{.}{%
}{.}\hspace{.4pt}1109\discretionary{/}{%
}{/}TVCG\hspace{.1pt}\discretionary{.}{%
}{.}\hspace{.4pt}2018\hspace{.1pt}\discretionary{.}{%
}{.}\hspace{.4pt}2871139}}


\bibitem{Liu2018Somewhere}
Y.~Liu and J.~Heer.
\newblock Somewhere over the rainbow: An empirical assessment of quantitative
  colormaps.
\newblock In {\em Proceedings of the 2018 CHI Conference on Human Factors in
  Computing Systems}, CHI~'18, pp. 598:1--598:12. ACM, 2018. doi: {{%
10\hspace{.1pt}\discretionary{.}{%
}{.}\hspace{.4pt}1145\discretionary{/}{%
}{/}3173574\hspace{.1pt}\discretionary{.}{%
}{.}\hspace{.4pt}3174172}}


\bibitem{Ma2020Explaining}
Y.~Ma, T.~Xie, J.~Li, and R.~Maciejewski.
\newblock Explaining vulnerabilities to adversarial machine learning through
  visual analytics.
\newblock {\em IEEE Transactions on Visualization and Computer Graphics},
  26(1):1075--1085, Jan. 2020. doi: {{%
10\hspace{.1pt}\discretionary{.}{%
}{.}\hspace{.4pt}1109\discretionary{/}{%
}{/}TVCG\hspace{.1pt}\discretionary{.}{%
}{.}\hspace{.4pt}2019\hspace{.1pt}\discretionary{.}{%
}{.}\hspace{.4pt}2934631}}


\bibitem{McInnes2018UMAP}
L.~{McInnes}, J.~{Healy}, and J.~{Melville}.
\newblock {UMAP}: Uniform manifold approximation and projection for dimension
  reduction.
\newblock {\em ArXiv e-prints}, 1802.03426, Feb. 2018.

\bibitem{MetaStackVisCode}
{MetaStackVis} code.
\newblock \url{http://bit.ly/MetaStackVis-code}, 2022.
\newblock Accessed December 10, 2022.

\bibitem{Moulavi2014Density}
D.~Moulavi, P.~A. Jaskowiak, R.~J. G.~B. Campello, A.~Zimek, and J.~Sander.
\newblock {\em Density-based clustering validation}, pp. 839--847.
\newblock 2014. doi: {{%
10\hspace{.1pt}\discretionary{.}{%
}{.}\hspace{.4pt}1137\discretionary{/}{%
}{/}1\hspace{.1pt}\discretionary{.}{%
}{.}\hspace{.4pt}9781611973440\hspace{.1pt}\discretionary{.}{%
}{.}\hspace{.4pt}96}}


\bibitem{Naimi2018Stacked}
A.~I. Naimi and L.~B. Balzer.
\newblock Stacked generalization: {A}n introduction to super learning.
\newblock {\em European Journal of Epidemiology}, 33(5):459--464, 2018.

\bibitem{Pedregosa2011Scikit}
F.~Pedregosa, G.~Varoquaux, A.~Gramfort, V.~Michel, B.~Thirion, O.~Grisel,
  M.~Blondel, P.~Prettenhofer, R.~Weiss, V.~Dubourg, J.~Vanderplas, A.~Passos,
  D.~Cournapeau, M.~Brucher, M.~Perrot, and E.~Duchesnay.
\newblock Scikit-learn: {M}achine learning in {P}ython.
\newblock {\em Journal of Machine Learning Research}, 12:2825--2830, Nov. 2011.
  doi: {{%
10\hspace{.1pt}\discretionary{.}{%
}{.}\hspace{.4pt}5555\discretionary{/}{%
}{/}1953048\hspace{.1pt}\discretionary{.}{%
}{.}\hspace{.4pt}2078195}}


\bibitem{plotly}
{Plotly} --- {Python} open source graphing library.
\newblock \url{https://plot.ly}, 2013.
\newblock Accessed December 10, 2022.

\bibitem{Sagi2018Ensemble}
O.~Sagi and L.~Rokach.
\newblock Ensemble learning: A survey.
\newblock {\em WIREs Data Mining and Knowledge Discovery}, 8(4):e1249,
  July--Aug. 2018. doi: {{%
10\hspace{.1pt}\discretionary{.}{%
}{.}\hspace{.4pt}1002\discretionary{/}{%
}{/}widm\hspace{.1pt}\discretionary{.}{%
}{.}\hspace{.4pt}1249}}


\bibitem{Shahhosseini2022Optimizing}
M.~Shahhosseini, G.~Hu, and H.~Pham.
\newblock Optimizing ensemble weights and hyperparameters of machine learning
  models for regression problems.
\newblock {\em Machine Learning with Applications}, 7:100251, 2022. doi: {{%
10\hspace{.1pt}\discretionary{.}{%
}{.}\hspace{.4pt}1016\discretionary{/}{%
}{/}j\hspace{.1pt}\discretionary{.}{%
}{.}\hspace{.4pt}mlwa\hspace{.1pt}\discretionary{.}{%
}{.}\hspace{.4pt}2022\hspace{.1pt}\discretionary{.}{%
}{.}\hspace{.4pt}100251}}


\bibitem{Smith1988Using}
J.~{Smith}, J.~{Everhart}, W.~{Dickson}, W.~{Knowler}, and R.~{Johannes}.
\newblock Using the {ADAP} learning algorithm to forecast the onset of diabetes
  mellitus.
\newblock In {\em Proceedings of the Annual Symposium Computer Application in
  Medical Care}, pp. 261--265. American Medical Informatics Association, 1988.

\bibitem{StackGenVisCode}
{StackGenVis} code.
\newblock \url{http://bit.ly/StackGenVis-code}, 2021.
\newblock Accessed December 10, 2022.

\bibitem{streamlit}
{Streamlit} --- {T}he fastest way to build and share data apps.
\newblock \url{https://streamlit.io}, 2020.
\newblock Accessed December 10, 2022.

\bibitem{Ting1997Stacked}
K.~M. Ting and I.~H. Witten.
\newblock Stacked generalization: When does it work?
\newblock In {\em Proceedings of the Fifteenth International Joint Conference
  on Artifical Intelligence --- Volume 2}, IJCAI~'97, pp. 866--871. Morgan
  Kaufmann Publishers Inc., 1997.

\bibitem{Wolpert1992Stacked}
D.~H. Wolpert.
\newblock Stacked generalization.
\newblock {\em Neural networks}, 5(2):241--259, 1992.

\bibitem{Xu2019EnsembleLens}
K.~Xu, M.~Xia, X.~Mu, Y.~Wang, and N.~Cao.
\newblock {EnsembleLens}: Ensemble-based visual exploration of anomaly
  detection algorithms with multidimensional data.
\newblock {\em IEEE Transactions on Visualization and Computer Graphics},
  25(1):109--119, Jan. 2019. doi: {{%
10\hspace{.1pt}\discretionary{.}{%
}{.}\hspace{.4pt}1109\discretionary{/}{%
}{/}TVCG\hspace{.1pt}\discretionary{.}{%
}{.}\hspace{.4pt}2018\hspace{.1pt}\discretionary{.}{%
}{.}\hspace{.4pt}2864825}}


\bibitem{Zhang2022SliceTeller}
X.~Zhang, J.~P. Ono, H.~Song, L.~Gou, K.-L. Ma, and L.~Ren.
\newblock {SliceTeller} : {A} data slice-driven approach for machine learning
  model validation.
\newblock {\em IEEE Transactions on Visualization and Computer Graphics}, pp.
  1--11, 2022. doi: {{%
10\hspace{.1pt}\discretionary{.}{%
}{.}\hspace{.4pt}1109\discretionary{/}{%
}{/}TVCG\hspace{.1pt}\discretionary{.}{%
}{.}\hspace{.4pt}2022\hspace{.1pt}\discretionary{.}{%
}{.}\hspace{.4pt}3209465}}


\end{thebibliography}
\end{document}